\documentclass[aps,pre,showkeys,groupedaddress]{revtex4}
\usepackage{amsfonts}
\usepackage{mathrsfs}
\usepackage{graphicx}
\usepackage{amsmath}
\usepackage{amssymb}
\usepackage{amsbsy}
\usepackage{bm}
\usepackage{algorithm}
\usepackage{algorithmic}
\usepackage{multirow}

\begin{document}

\title{Reinforced stochastic gradient descent for deep neural network learning}

\author{Haiping Huang}
\email{physhuang@gmail.com}
\affiliation{RIKEN Brain Science Institute, Wako-shi, Saitama
351-0198, Japan}
\author{Taro Toyoizumi}
\email{taro.toyoizumi@brain.riken.jp}
 \affiliation{RIKEN Brain
Science Institute, Wako-shi, Saitama 351-0198, Japan}
\date{\today}

\begin{abstract}
Stochastic gradient descent (SGD) is a standard optimization method to minimize a training error with respect to network parameters in modern neural network learning. However, it typically suffers from proliferation of saddle points
in the high-dimensional parameter space.
Therefore, it is highly desirable to design an efficient algorithm to escape from these saddle points and reach a parameter region of better generalization capabilities. Here, we propose a simple extension of SGD, namely reinforced SGD, which
simply adds previous first-order gradients in a stochastic manner with a probability that increases with learning time. As verified in a simple synthetic dataset, this method
significantly accelerates learning compared with the original SGD. Surprisingly, it dramatically reduces over-fitting effects, even compared with state-of-the-art
adaptive learning algorithm---Adam. For a benchmark handwritten digits dataset, the learning performance is comparable to Adam, yet with an
extra advantage of requiring one-fold less computer memory. The reinforced SGD is also compared with SGD with fixed or adaptive momentum parameter and Nesterov's momentum, which shows that the 
proposed framework is able to reach a similar generalization accuracy with less computational costs. Overall, our method introduces stochastic memory into gradients, which plays an important role in understanding how gradient-based training algorithms can work
and its relationship with generalization abilities of deep networks.

\end{abstract}
 \maketitle
\section{Introduction}
Multilayer neural networks have achieved state-of-the-art performances in image recognition~\cite{Hinton-2012imag}, speech recognition, and even natural
language processing~\cite{DL-2016}. This impressive success is based on a simple powerful stochastic gradient descent (SGD) algorithm~\cite{robbins-1951}, and its variants. 
This algorithm estimates gradients of an error function based on mini-batches of an entire dataset. Gradient noise caused by mini-batches helps exploration of parameter space to
some extent. However, the parameter space is highly non-convex for a typical deep network training, and finding a good path for SGD to improve generalization ability of deep neural
networks is thus challenging~\cite{Bengio-2010}. 

As found in standard spin glass models of neural networks~\cite{Huang-2014,Baldassi-2016}, a non-convex error surface is accompanied by exponentially many local minima, which hides the (isolated) global
minima and thus makes any local
search algorithms easily get trapped. In addition, the error surface structure of deep networks might behave similarly to random Gaussian error surface~\cite{Bray-2007,Fyo-2007}, which demonstrates that
critical points (defined as zero-gradient points) of high error have a large number of negative eigenvalues of the corresponding
Hessian matrix. Consistent with this theoretical study, empirical studies on deep network training~\cite{Lecun-2014,Dau-2014} showed that
SGD is slowed down by a proliferation of saddle points with many negative curvatures and even plateaus (eigenvalues close to zero in many directions). The prevalence of
saddle points poses an obstacle to attain better generalization properties for a deep network, especially for SGD based on first-order optimization, while second-order optimization relying on
Hessian-vector products is more computationally expensive~\cite{Ge-2015}. The second order method that relies on positive-definite curvature approximations, can not follow directions
with negative curvature, and is easily trapped by saddle points~\cite{Ba-2017}.

In this paper, we show a heuristic strategy to overcome the plateaus problem for SGD learning. We call this strategy reinforced SGD (R-SGD), which provides
a new effective strategy to use the gradient information, i.e., to update one network parameter, an instantaneous gradient is reinforced by (accumulated) previous gradients with an 
increasing reinforcement probability that grows with learning time steps. In other words, the reinforcement may be turned off, and then only the instantaneous gradient is used for learning. 
This kind of stochastic reinforcement enhances the exploration of parameter space. The excellent performance of R-SGD is
verified first on training a toy fully-connected deep network model to learn a simple non-linear mapping generated by a two-layer feedforward network, and then on a benchmark handwritten digits dataset~\cite{Lecun-1998}, 
in comparison to both vanilla backpropagation (BackProp)~\cite{Back-1988} and state-of-the-art Adam algorithm~\cite{Adam}. In the benchmark dataset, we also clarify the performance difference between R-SGD and SGD with 
fixed or adaptive momentum parameter~\cite{Suts-2013} and Nesterov's momentum~\cite{nag-2013}.

\section{Fully-connected deep networks}
\label{fcnn}
We consider a toy deep network model with $L$ layers of fully-connected feedforward architecture. 
Each layer has $n^k$ neurons (so-called width of layer $k$). We define the input as $n^1$-dimensional vector $\mathbf{v}$, 
and the weight matrix $\mathbf{W}^k$ specifies the symmetric connections between layer $k$ and layer $k-1$. The symmetry means that the same connections are used to backpropagate the error during
training.
A bias parameter can also be incorporated into the weight matrix by assuming an additional constant input. 
The output at the final layer is expressed as:
\begin{equation}\label{output}
    \mathbf{y}=f_L\left(\mathbf{W}^Lf_{L-1}(\mathbf{W}^{L-1}\cdots f_2(\mathbf{W^2}\mathbf{v}))\right),
\end{equation}
where $f_k(\cdot)$ is an element-wise sigmoid function for neurons at layer $k$, defined as $f(x)=\frac{1}{1+e^{-x}}$, unless otherwise specified (e.g., ReLU activation funtion).
The network is trained to 
learn the target mapping generated randomly as $\{\mathbf{v}^m,\mathbf{y}_{*}^m\}_{m=1}^{M}$, where the input is generated 
from a standard normal distribution with zero mean and unit variance, and the target label $y_{*}$ is generated according to 
the non-linear mapping $y_{*}=f(\mathbf{W}_{g}\mathbf{v})$, in which each entry of the data-generating matrix $\mathbf{W}_{g}$ 
follows independently a standard normal distribution as well. The deep network is trained to learn this non-linear mapping (continuous target labels) from a set of examples. 
We generate a total of $2M$ examples, in which the first $M$ examples are used for training and the last $M$ examples are used for testing
to evaluate the generalization ability of the learned model.

In simulations, we use deep network architecture of $L=4$ layers to learn the target non-linear mapping, in which the network is thus specified by $n^{1}$-$n^{2}$-$n^{3}$-$n^{4}$, 
with $n^1$ indicating the dimension of the input data and $n^L$ the dimension of the output. We use this simple toy setting to test
our idea first, and then the idea is further verified in the handwritten digits dataset.

\section{Backpropagation and its variants}
\label{R-BP}
We first introduce the vanilla BackProp~\cite{Back-1988} for training the deep network defined in Sec.~\ref{fcnn}. 
We use quadratic loss (error) function defined as $E=\frac{1}{2}\boldsymbol{\epsilon}^{{\rm T}}\boldsymbol{\epsilon}$, where $^{\rm T}$ denotes a vector (matrix) transpose operation, and $\boldsymbol{\epsilon}$ defines 
the difference between the target and actual outputs as $\boldsymbol{\epsilon}=\mathbf{y}_{*}-\mathbf{y}$. To backpropagate the error, 
we also define two associated quantities: one is the state of neurons at $k$-th layer defined by $\mathbf{s}^k$ (e.g., $\mathbf{s}^L=\mathbf{y}$, $\mathbf{s}^{1}=\mathbf{v}$), and the other is the weighted-sum input 
to neurons at $k$-th layer defined by $\mathbf{h}^k\equiv\mathbf{W}^{k}\mathbf{s}^{k-1}$. Accordingly, we define two related gradient vectors:
\begin{subequations}\label{grad}
\begin{align}
    \boldsymbol{\delta}^k&\equiv\frac{\partial E}{\partial\mathbf{s}^k},\\
\boldsymbol{\kappa}^k&\equiv\frac{\partial E}{\partial\mathbf{h}^k},
\end{align}
\end{subequations}
which will be used to derive the propagation equation based on the chain rule. It is straightforward to 
derive $\boldsymbol{\kappa}^L=-\boldsymbol{\epsilon}\circ\mathbf{y}'$, where $\circ$ indicates the element-wise multiplication, and $\mathbf{y}'$ is the derivative
of the non-linear transfer function with respect to its argument.
By applying the chain rule, we obtain the weight update equation for the top layer as
\begin{equation}\label{top}
    \Delta \mathbf{W}^L=-\eta\boldsymbol{\kappa}^L(\mathbf{s}^{L-1})^{{\rm T}},
\end{equation}
where $\eta$ is the learning rate, and the remaining part is the gradient information, which indicates
how a small perturbation to the weight affects the change of the error computed at the top (output) layer.

To update the weight parameters at lower layers, we first derive the propagating equations for gradient vectors as follows:
\begin{subequations}\label{backprop}
\begin{align}
    \boldsymbol{\delta}^k&=(\mathbf{W}^{k+1})^{{\rm T}}\boldsymbol{\kappa}^{k+1},\\
\boldsymbol{\kappa}^{k}&=\boldsymbol{\delta}^{k}\circ(\mathbf{f}_{k})',
\end{align}
\end{subequations}
where $k\leq L-1$.
Using the above backpropagation equation, the weight at lower layers is updated as:
\begin{equation}\label{lower}
    \Delta\mathbf{W}^{k}=-\eta\boldsymbol{\kappa}^{k}(\mathbf{s}^{k-1})^{{\rm T}},
\end{equation}
where $k\leq L-1$.
The neural state used to update the weight parameters comes from a forward pass from the input vector to the output vector at the top layer.
A forward pass combined with a backward propagation of the error forms the vanilla BackProp widely used in training deep networks given the labeled data~\cite{Stra-2009}. 
To improve the training efficiency, one usually divides the entire large dataset into a set of mini-batches, each of which is used to get the average gradients across the examples within that mini-batch.
One epoch corresponds to a sweep of the full dataset. The learning time is thus measured in units of epochs. For one epoch, the weight is actually updated for $M/B$ times ($B$ is the size of a mini-batch).
This process is usually termed SGD. 

Here, we briefly introduce two kinds of SGD with momentum techniques. The first one is the SGD with momentum (SGDM). The learning equation is revised as
\begin{subequations}\label{sgdm-eq}
\begin{align}
    \boldsymbol{\nu}_{t}&=\rho_t\boldsymbol{\nu}_{t-1}+\mathbf{g}_{t},\\
\Delta\mathbf{W}_t&=-\eta\boldsymbol{\nu}_t,
\end{align}
\end{subequations}
where $\mathbf{g}_t\equiv\nabla_{\mathbf{W}}E(\mathbf{W}_{t-1})$ denotes the gradient estimated from the average over examples within the current mini-batch,
and $\rho_t$ is the momentum
parameter, which can be either prefixed ($\rho_t=\rho$ in the classical SGDM) or varied over learning steps ($t$).

The second one is Nesterov's accelerated gradient (NAG)~\cite{nag-2013}, which first implements a partial update to $\mathbf{W}_t$, and then uses the updated $\mathbf{W}_t$ to evaluate gradients, i.e.,
\begin{subequations}\label{nag-eq}
\begin{align}
    \boldsymbol{\nu}_{t}&=\rho_t\boldsymbol{\nu}_{t-1}-\eta\mathbf{g}'_{t},\\
\Delta\mathbf{W}_t&=\boldsymbol{\nu}_t,
\end{align}
\end{subequations}
where $\mathbf{g}'_{t}\equiv\nabla_{\mathbf{W}}E(\mathbf{W}_{t-1}+\rho_t\boldsymbol{\nu}_{t-1})$. Note that the partial update takes an extra computational cost of $T_{max}\frac{M}{B}|\mathbf{W}|$, where $T_{max}$ denotes 
the maximal number of epochs, and $|\mathbf{W}|$ denotes the total amount of network parameters.

\section{Reinforced stochastic gradient descent}
In the above vanilla BackProp, only current gradients are used to update the weight matrix. Therefore in a non-convex optimization, the backpropation gets easily stalled by the plateaus or
saddle points on the error surface, and it is hard to escape from these regions. During training, gradients may be very noisy with large fluctuations. 
If update directions along some weight components are stochastically allowed to accumulate the history of
gradient information, while other directions still follow the current gradients, the learning performance may be boosted. This stochastic rule of turning on accumulation may help
SGD to handle the uncertainty of updating the weights. We will test this idea in the following deep neural network learning.

To enable SGD to use previous gradient information, we define a stochastic process for updating modified gradient $\tilde{\mathbf{g}}_t$ used at each learning step as follows:
\begin{equation}\label{RBP}
(\tilde{\mathbf{g}}_t)_i\leftarrow\begin{cases}
                               (\mathbf{g}_t)_i, & \text{with prob. $1-\Gamma(t)$},\\
                               (\mathbf{g}_t)_i+(\tilde{\mathbf{g}}_{t-1})_i, &\text{with prob. $\Gamma(t)$}.
                              \end{cases}
\end{equation}
where the stochastic reinforcement is independently applied to each weight component, and $\tilde{\mathbf{g}}_{t-1}$ contains information
about the history of the evolving gradients, and the current gradients are reinforced by the previous accumulated gradients with a reinforcement probability defined by
$\Gamma(t)$. The stochastic rule in Eq.~(\ref{RBP}) is a switch-like (all-or-none) event; its smooth averaged version given $\tilde{\mathbf{g}}_{t-1}$ is $\mathbb{E}[\tilde{\mathbf{g}}_t|\tilde{\mathbf{g}}_{t-1}]=\mathbf{g}_t+\Gamma(t)\tilde{\mathbf{g}}_{t-1}$, where
$\Gamma(t)$ is equivalent to $\rho_t$ in SGDM with time-dependent momentum parameter. 
Using adaptive momentum parameter is important in 
boosting the learning performance of SGDM. However, the switch-like property is able to reach a better or equivalent test accuracy with fewer training steps.
Comparisons will be made on the handwritten digits dataset.

\begin{algorithm}
   \caption{R-SGD ($\eta_0,\beta,\gamma_0,\lambda$)}
   \label{alg:rbp}
\begin{algorithmic}
 \STATE {\bfseries Input:} data $\{\mathbf{v}^{m},\mathbf{y}_*^{m}\}_{m=1}^M$, $T_{max}=100$, mini-batch size $B$
   \STATE $t\leftarrow 0$
   \STATE $\gamma\leftarrow\gamma_0$
   \STATE $\eta\leftarrow\eta_0$
   \FOR{$t_{{\rm ep}}=1$ {\bfseries to} $T_{max}$}
    \FOR{$l=1$ {\bfseries to} $M/B$}
 \STATE $\mathbf{g}_t\leftarrow\frac{1}{B}\sum_{m=1}^{B}\nabla_{\mathbf{W}_{t-1}}E(\mathbf{v}^m,\mathbf{y}_{*}^{m})$
 \STATE 
         $\tilde{\mathbf{g}}_t\leftarrow$ Eq.~(\ref{RBP}) using $\gamma$\\
   \STATE       $\mathbf{W}_t\leftarrow\mathbf{W}_{t-1}-\eta\tilde{\mathbf{g}}_t$\\
        \STATE $t\leftarrow t+1$
\ENDFOR
\STATE $\gamma\leftarrow\gamma_0e^{-\lambda t_{{\rm ep}}}$
\STATE $\eta\leftarrow\eta\beta^{t_{{\rm ep}}}$
\ENDFOR
\end{algorithmic}
\end{algorithm}

\begin{figure}
\centering
 \includegraphics[bb=0 0 339 244,scale=0.7]{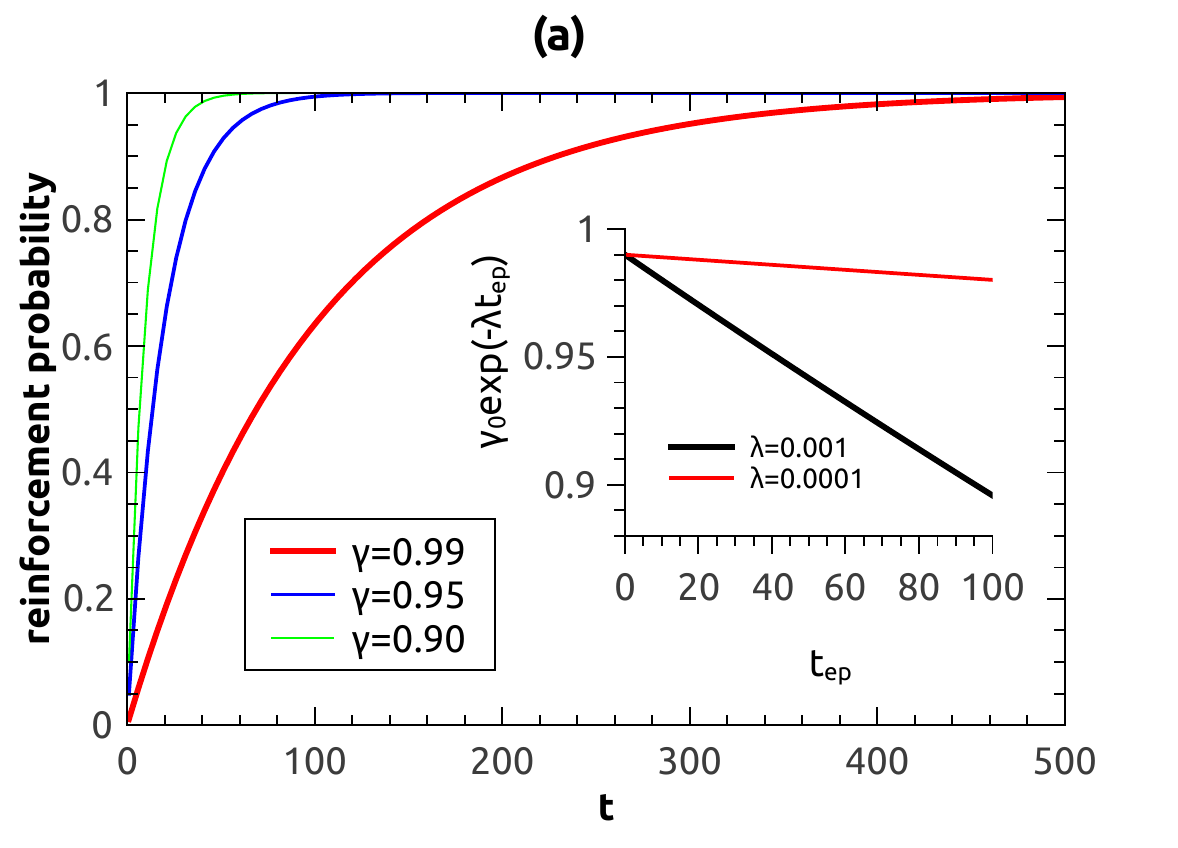}
   \hskip .05cm
   \includegraphics[bb=0 0 312 244,scale=0.7]{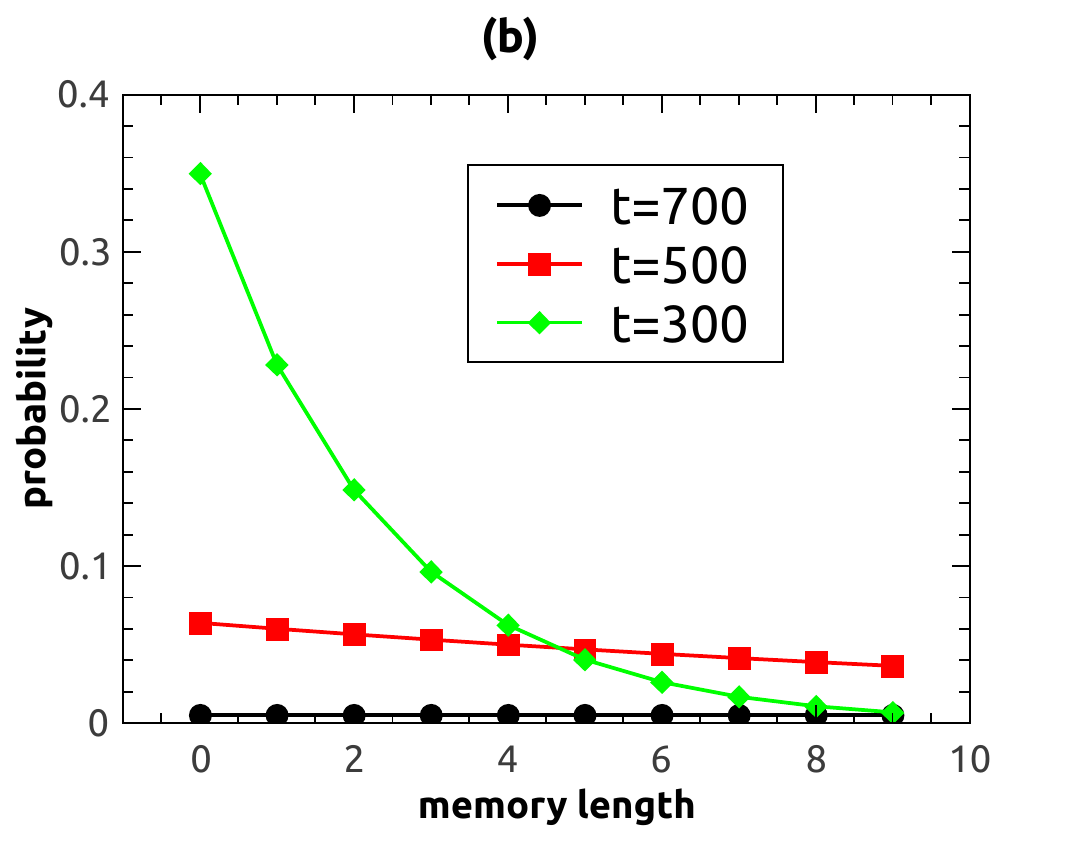}
  \caption{ (Color online) Properties of R-SGD. (a) Typical behavior of the reinforcement probability ($\Gamma(t)=1-\gamma^t$) as a function of iterations. The inset shows an exponential decay of $\gamma$ ($\gamma_0=0.99$).
  Note that the learning time is measured in units of epochs in the inset. (b) Probability distribution of memory length for R-SGD at different time-steps ($t=300,500$, and $700$). $(\gamma_0,\lambda)=(0.9995,0.0001)$.
  }\label{reprob}
\end{figure}

Eq.~(\ref{RBP}) is a very simple way to re-use the previous gradient information, and forms the key component of R-SGD.  We first choose $\Gamma(t)=1-\gamma^t$, where $\gamma=\gamma_0e^{-\lambda t_{{\rm ep}}}$. $\gamma_0$ and $\lambda$ are prefixed constants, and 
$t_{{\rm ep}}$ refers to the learning time in units of epochs. $\Gamma(t)$ can be rewritten as $1-e^{-t/\tau_{{\rm R}}}$, where $\tau_{{\rm R}}\equiv-\frac{1}{\ln\gamma_0-\lambda t_{{\rm ep}}}$ setting the time scale of the dynamics of the reinforcement probability.
$\gamma_0$ is usually fixed to a value very close to one, and $\lambda$ takes a small value. 
Therefore the current gradient has an increasing probability to be reinforced by the previous gradients, and retains its instantaneous value otherwise. This reinforcement probability is not 
the unique choice, e.g., $1-a_0/(t+1)^{b_0}$ ($a_0$ and $b_0$ are constants) is also a candidate (discussed in Sec.~\ref{res-mnist}).
 We show a typical trace of the reinforcement probability and $\gamma$ in Fig.~\ref{reprob} (a), 
and will test effects of hyper-parameters $(\gamma_0,\lambda)$ on training dynamics in Sec.~\ref{res-toy}. Note that by setting $(\gamma_0,\lambda)=(1,0)$, one recovers the vanilla BackProp.
In all simulations, we use an exponentially-decaying learning rate $\eta_{t_{{\rm ep}}}=\eta_{t_{{\rm ep}}-1}\beta^{t_{{\rm ep}}}$, where
$\eta_0=0.8$ and $\beta=0.999$, with a minimal learning rate of $0.02$, unless otherwise specified. R-SGD is summarized in algorithm~\ref{alg:rbp}.

The gradient used in R-SGD may be the accumulated one (over an unfixed or stochastic number
  of consecutive steps), which
  contains short or long-term memory of previous gradient information (Fig.~\ref{reprob} (b)). Hence, the step-size is
  a sum of previous gradients over a memory length $\mathcal{L}$, which follows a probability $P_t(\mathcal{L})$ decaying with $\mathcal{L}$ (Fig.~\ref{reprob} (b)). This stochastic process is summarized by
  \begin{subequations}\label{rbplen}
\begin{align}
    \Delta \mathbf{W}_t^{{\rm R-SGD}}&=-\eta_t\sum_{l=t-\mathcal{L}}^{t}\mathbf{g}_l,\\
\mathcal{L}&\sim P_t(\mathcal{L})=(1-\Gamma(t-\mathcal{L}))\exp\left(\sum_{l=t-\mathcal{L}+1}^{t}\ln\Gamma(l)\right),
\end{align}
\end{subequations}
where the prefactor indicates the probability that the memory is cleaned before accumulation. This probability is normalized since $\sum_{\mathcal{L}=0}^{t}P_t(\mathcal{L})=1-\prod_{l=0}^t\Gamma(l)=1$, where
$\Gamma(0)=0$. In the SGDM, the momentum term is deterministically added to the learning step size with coefficient $\rho_t$ (Eq.~(\ref{sgdm-eq})). Unfolding this process, we obtain the step-size as
 \begin{equation}\label{momen}
    \Delta \mathbf{W}_t^{{\rm SGDM}}=-\eta_t\sum_{l=1}^{t}\Bigl(\prod_{l'=l+1:l\neq t}^t\rho_{l'}+\delta_{l,t}\Bigr)\mathbf{g}_l,
\end{equation}
where $\delta_{l,t}$ is a Kronecker delta function, and each gradient is weighted by a value smaller than one. 

To show the efficiency of R-SGD, we also compare its performance with that of a state-of-the-art stochastic optimization algorithm, namely adaptive moment estimation (Adam)~\cite{Adam}.
Adam performs a running average of gradients and their second raw moments, which are used to adaptively change the learning step-size. We use heuristic 
parameters of Adam given in~\cite{Adam}, except that $\eta_0=0.01$ with the lowest value set to $0.001$. Large $\eta_0$ as we use in R-SGD does not work
in our simulations for Adam.

\section{Results and discussion}
\label{res}
\subsection{Learning performance in simple synthetic dataset}
\label{res-toy}
\begin{figure}
\centering
 \includegraphics[bb=0 0 382 306,scale=0.6]{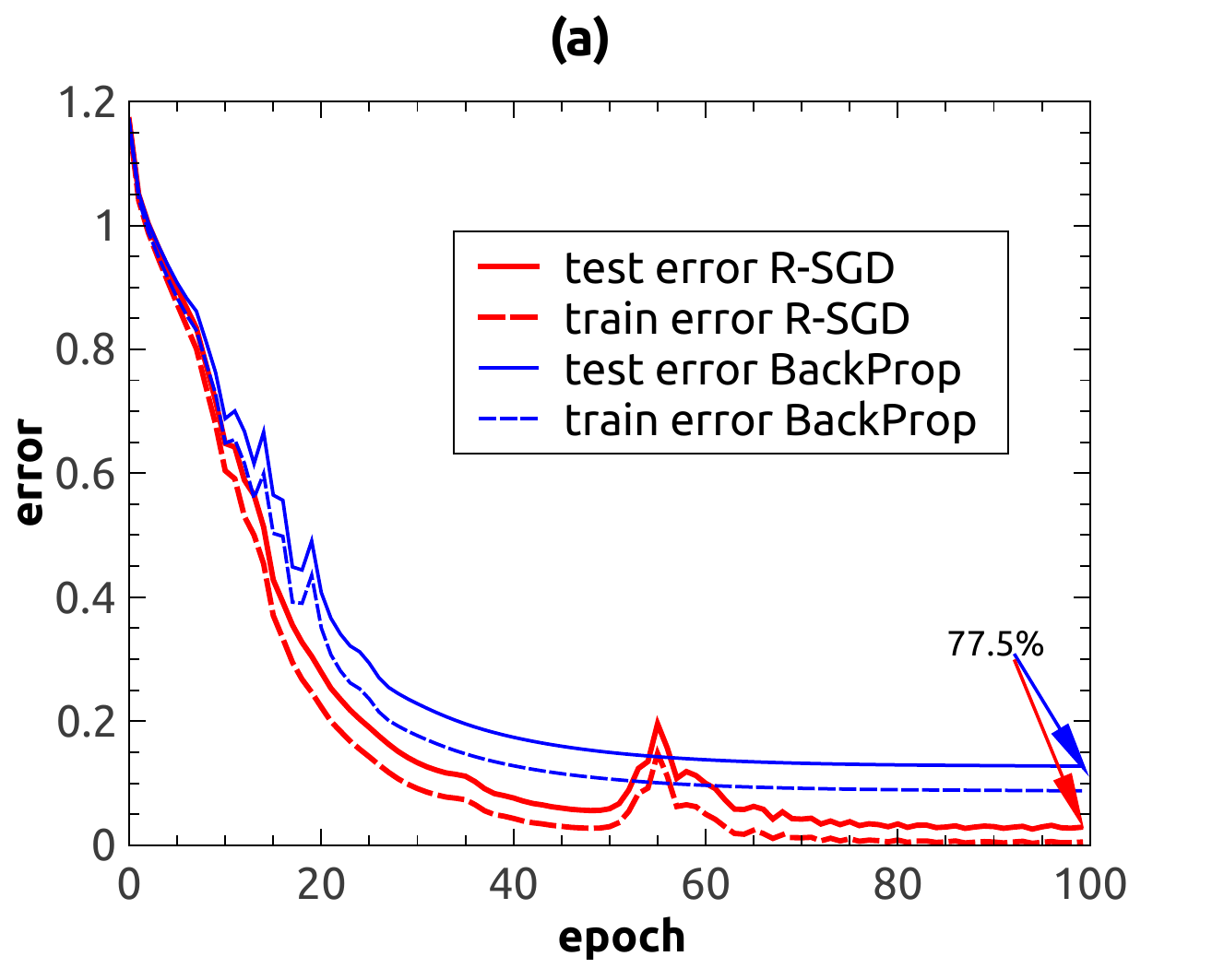}
     \hskip .05cm
  \includegraphics[bb=0 0 443 345,scale=0.55]{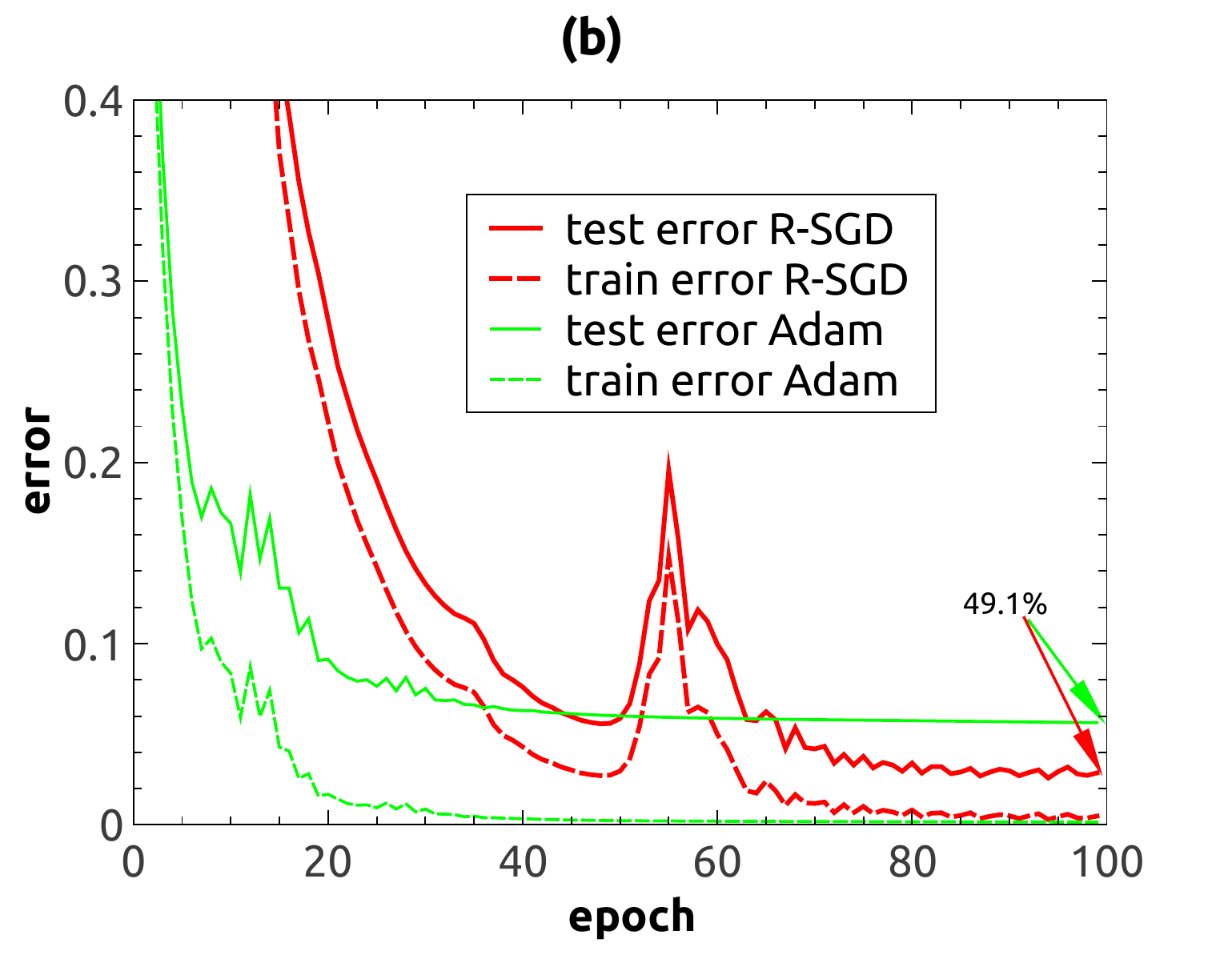}
     \vskip .05cm
  \caption{(Color online) Test performance of R-SGD compared with those obtained by using the vanilla BackProp (a) and Adam (b), based on $M=1000$ training examples. Note that the test performance is
  improved by $77.5\%$, compared with BackProp, and by $49.1\%$, compared with Adam.
     }\label{comprbp}
 \end{figure}

We first test our idea in the simple synthetic dataset described in Sec.~\ref{fcnn}. We use a $4$-layer deep network architecture as $100$-$400$-$200$-$10$. Training examples are divided into
mini-batches of size $B=100$. In simulations, we use the parameters 
$(\gamma_0,\lambda)=(0.9995,0.0001)$, unless otherwise specified. Although the chosen parameters are not optimal to achieve the best performance, we still observe the outstanding performance of R-SGD.
In Fig.~\ref{comprbp} (a), we compare the vanilla BackProp with R-SGD. We clearly see that the test performance is finally improved at $100$-th
epoch by a significant amount (about $77.5\%$). Meanwhile, the training error is also significantly lower than that of BackProp. A salient feature of R-SGD is that,
at the intermediate stage, the reinforcement strategy guides SGD to escape from possible plateau regions of high error surrounding saddle points, and finally reach a region of
very nice generalization properties. This process is indicated by the temporary peak in both training and test errors for R-SGD. Remarkably, even before or after this peak, there are a few less
significant fluctuations in both training and test errors. These fluctuations play an important role in the exploration of the parameter space.

Compared to state-of-the-art Adam, R-SGD still improves the final test performance by a significant
amount (about $49.1\%$, see Fig.~\ref{comprbp} (b)). Note that Adam is able to decrease both training and test errors very quickly, but the decrease becomes
slow after about $40$ epochs. In contrast, R-SGD keeps decreasing both errors by a more significant amount than Adam, despite the presence of slightly significant fluctuations. Another key feature of Fig.~\ref{comprbp} (b) is that, a region in the parameter space with low training error does not generally have low test error. The 
training error reached by R-SGD is clearly higher than that of Adam, but the network architecture learned by R-SGD has nicer generalization property. This observation is
consistent with a recent study of maximizing local entropy in deep networks~\cite{EGD-2016}.

\begin{figure}
\centering
    \includegraphics[bb=0 0 444 361,scale=0.5]{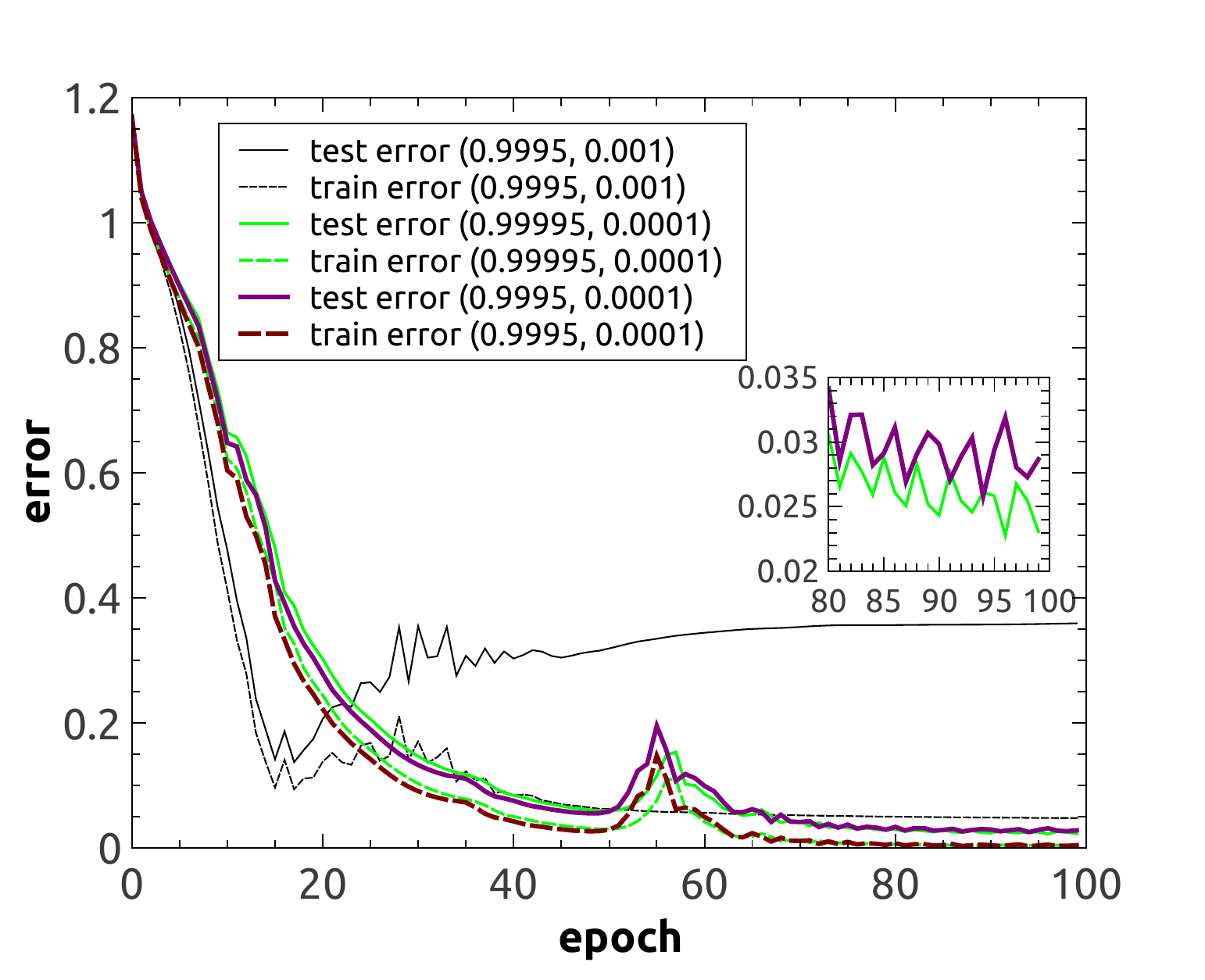}
  \caption{
  (Color online) Effects of reinforcement parameters ($\gamma_0$,$\lambda$) on the learning performance for R-SGD based on $M=1000$ training examples. The inset is an enlarged view for
  the later stage (the worst performance with ($0.9995$,$0.001$) is omitted for comparison).
  }\label{Epara}
\end{figure}


We then study the effects of reinforcement parameters $(\gamma_0,\lambda)$ on the learning performance, as shown in Fig.~\ref{Epara}.
If the exponential decay rate $\lambda$ is large, R-SGD over-fits the data rapidly at around $17$ epochs. This is because, $\gamma$
decays rapidly from $\gamma_0$, and thus a stochastic fluctuation at earlier stages of learning is strongly suppressed, which limits severely
the exploration ability of R-SGD in the high-dimensional parameter space. In this case, R-SGD is prone to get stuck by bad
regions with poor generalization performances. However, maintaining the identical small decay rate, we find that a larger value of $\gamma_0$ leads to a smaller test error (inset of Fig.~\ref{Epara}). For relatively large values of
$\gamma_0$, the learning performance is not radically different. 

\begin{figure}
\centering
    \includegraphics[bb=0 0 325 244,scale=0.7]{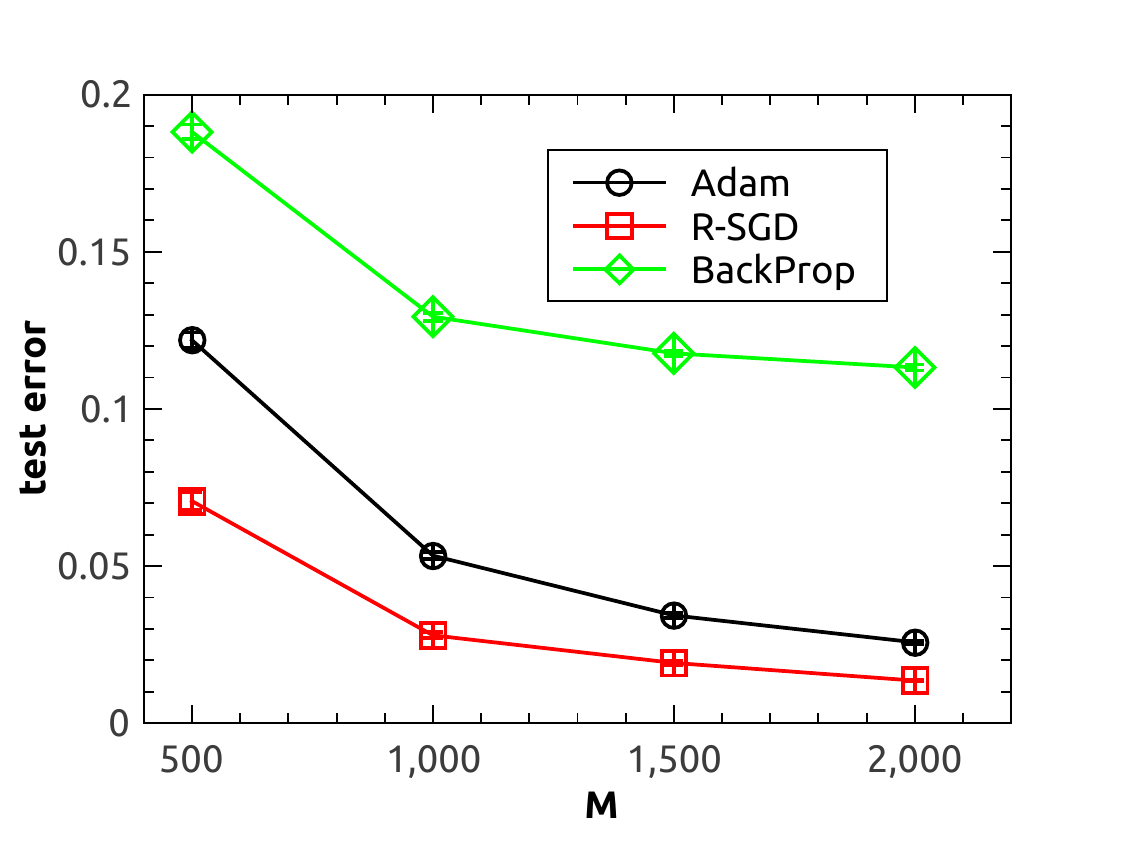}
  \caption{
  (Color online) Data size dependence of test performance for R-SGD, BackProp and Adam. The result is averaged over ten independent runs.
  }\label{Esamp}
\end{figure}
We also study the effects of training data size ($M$) on the learning performances.
Clearly, we see from Fig.~\ref{Esamp}, the test error decreases with the training data size as expected. R-SGD outperforms the
vanilla BackProp, and even Adam. 

For the simple toy model, SGDM with fixed momentum parameter could outperform
  Adam with a careful optimization of momentum parameter (e.g., $\rho_t=0.9$ $\forall t$), but R-SGD still outperforms the classical SGDM (fixed momentum parameter) by about $14.3\%$ when $M=1000$. By adaptively changing the momentum parameter whose value is the same as
  the reinforcement probability of R-SGD (i.e., a smooth averaged version of R-SGD),
  SGDM can reach a similar performance to that of R-SGD. Because the synthetic data in the toy model is relatively simple and the reinforcement
  probability used in the synthetic data is rapidly saturated to one (Fig.~\ref{reprob} (a)), it is difficult to show the performance difference between R-SGD and SGDM with the adaptive momentum parameter. Therefore, for MNIST classification
  task in the next section, we use the reinforcement probability of $\Gamma(t)=1-1/\sqrt{t+1}$, which does not rapidly approach one, and compare R-SGD with different variants of momentum-based SGD.

\begin{figure}
\centering
 \includegraphics[bb=0 0 312 244,scale=0.7]{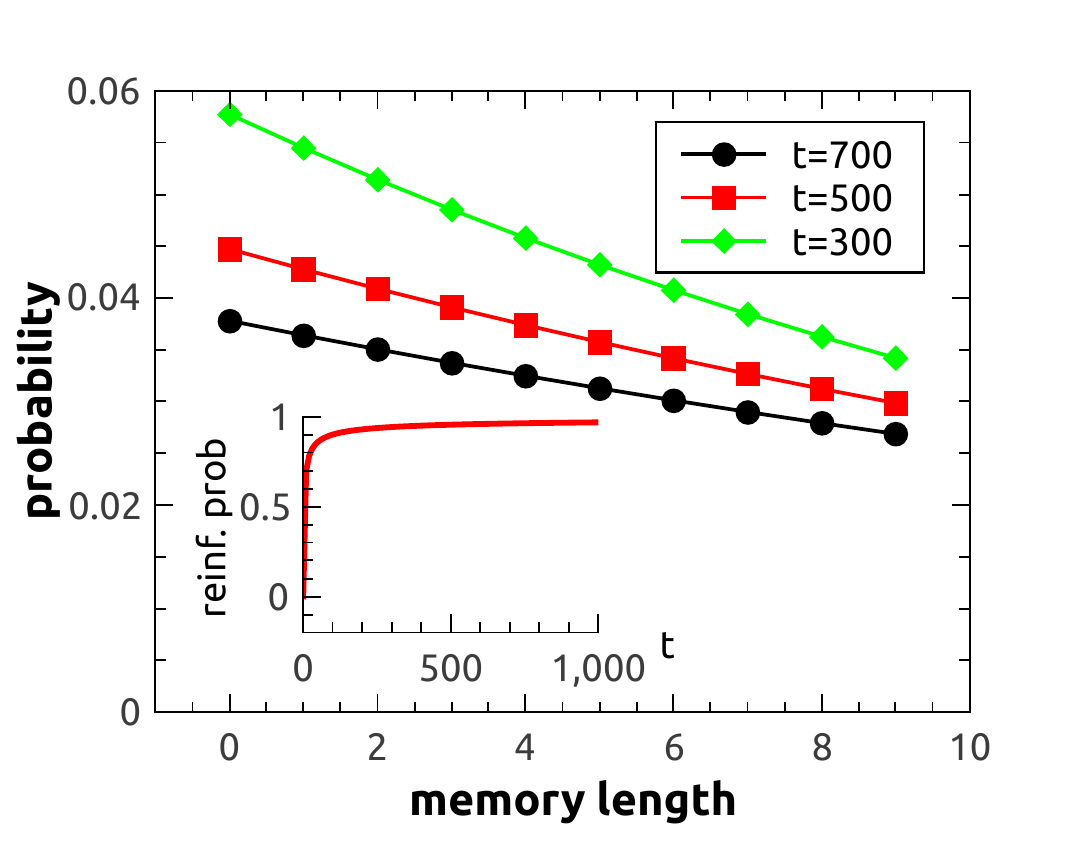}
  \caption{ (Color online) Probability distribution of memory length ($P_t(\mathcal{L})$) for R-SGD at different time-steps ($t=300,500,$ and $700$) for a reinforcement probability $\Gamma(t)=1-1/\sqrt{t+1}$ (inset) we used in learning
  MNIST dataset.
  }\label{powerlaw}
\end{figure}
\subsection{Learning performance in MNIST dataset}
\label{res-mnist}
Finally, we evaluate the test performance of R-SGD on MNIST dataset. The MNIST handwritten digits dataset contains
$60000$ training images and an extra $10000$ images for testing. Each image is one of ten handwritten digits ($0$ to $9$) with $28\times28$ pixels.
Therefore the input dimension is $n^1=784$. For simplicity, we choose the network structure as $784$-$100$-$200$-$10$.

\begin{figure}
\centering
    \includegraphics[bb=0 0 472 390,scale=0.6]{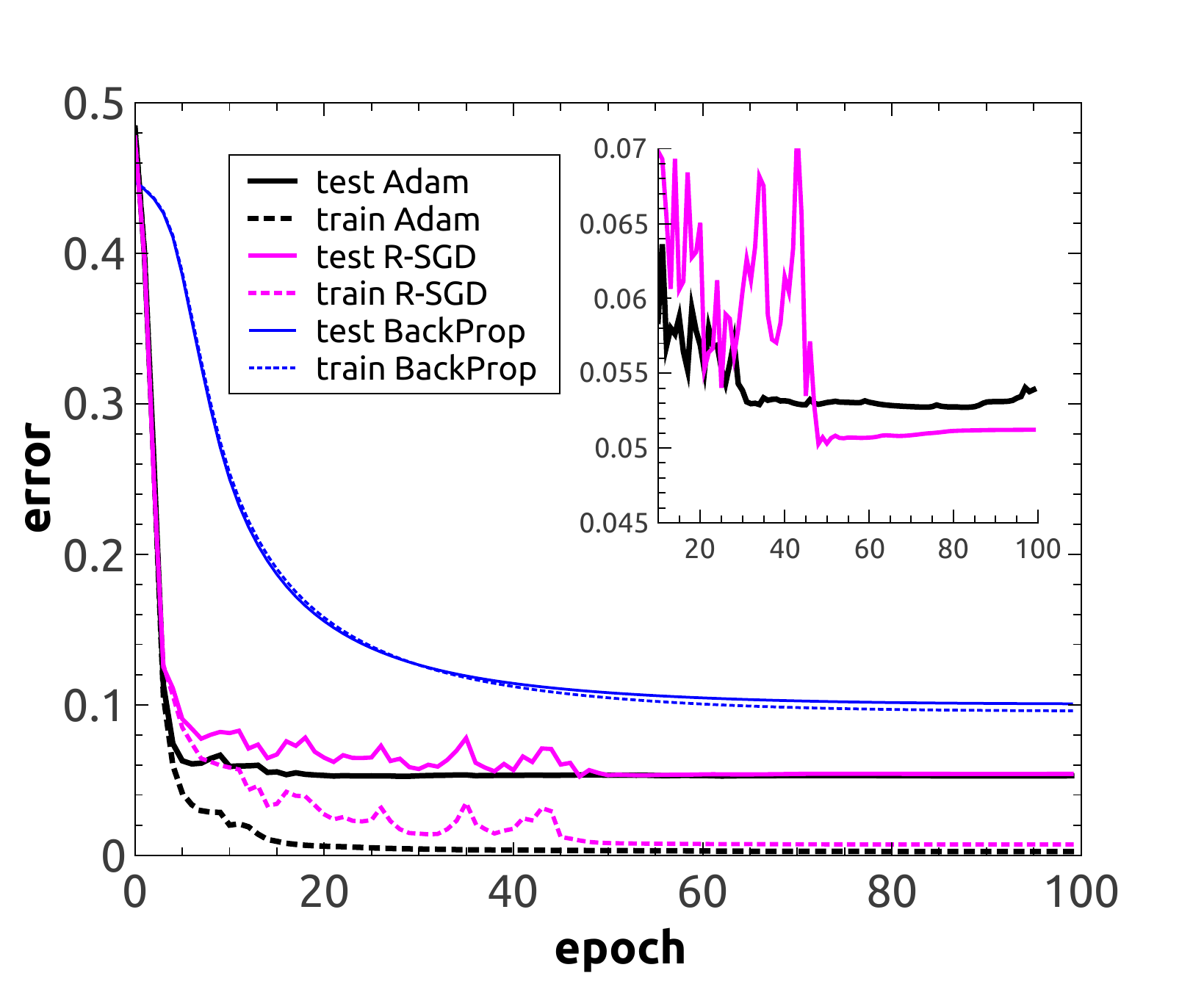}
  \caption{
  (Color online) Learning performance of R-SGD on MNIST dataset based on $M=10000$ training examples with $2000$ testing images, compared with BackProp and Adam. Mini-batch size $B=250$.
  The inset shows a training trajectory (test error) when $M=15000$, suggesting that Adam can be trapped by a higher plateau.
  }\label{mnist}
\end{figure}

\begin{figure}
\centering
   \includegraphics[bb=0 0 325 243,scale=0.7]{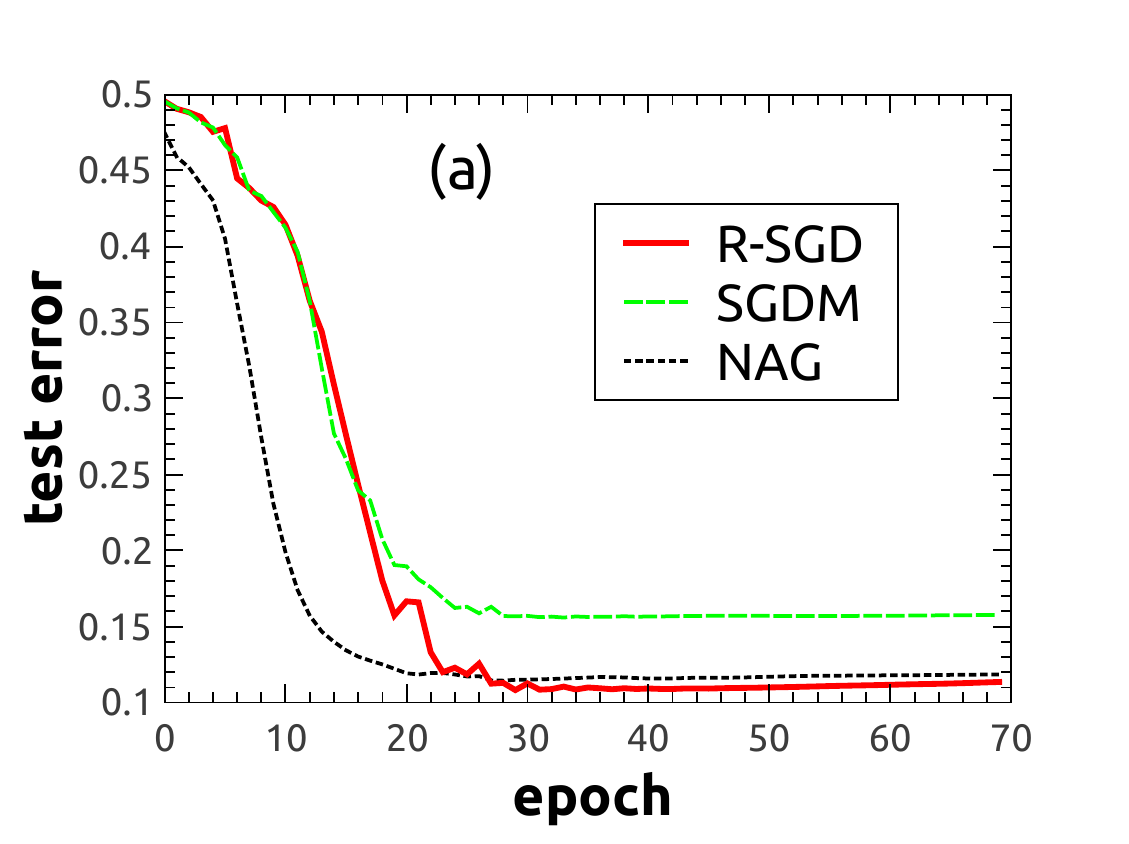}
     \hskip .05cm
      \includegraphics[bb=0 0 325 244,scale=0.7]{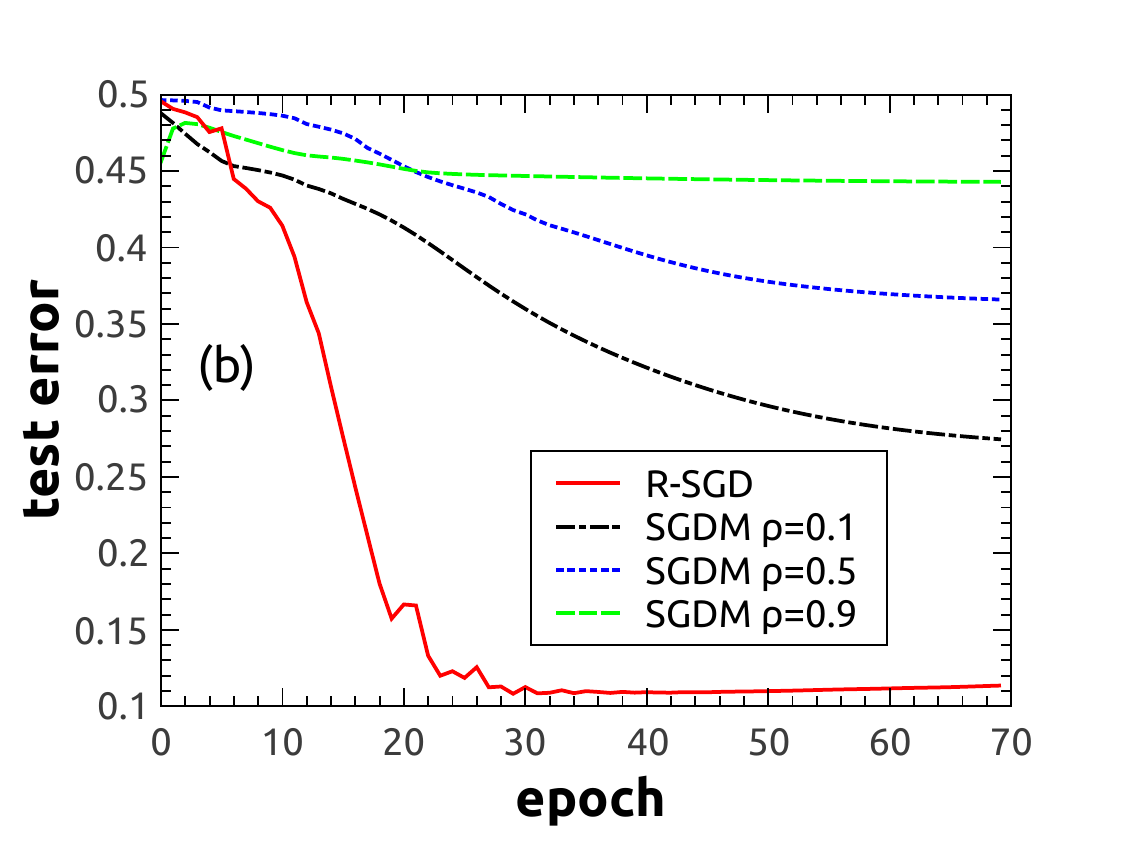}
      \vskip .05cm
       \includegraphics[bb=0 0 325 244,scale=0.7]{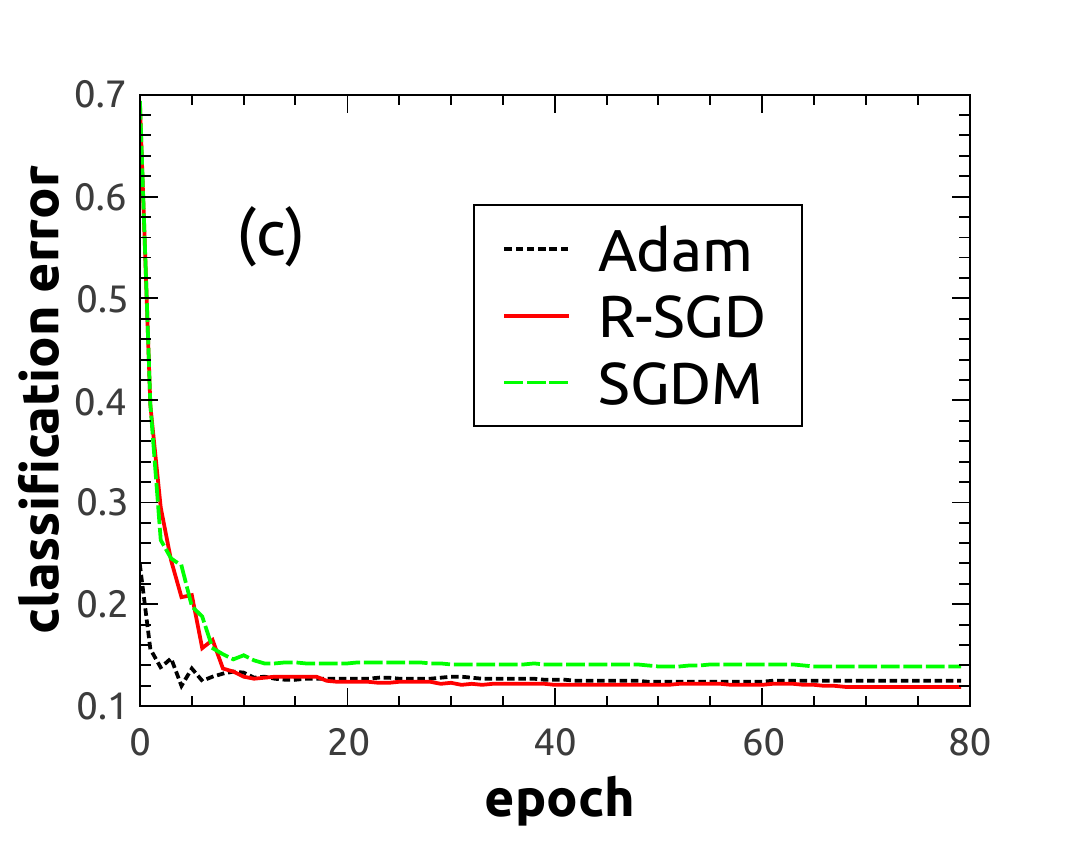}
  \caption{
  (Color online)  Comparison of R-SGD with variants of momentum-based SGD on MNIST dataset ($M=1000$ training examples and $1000$ testing examples). Mini-batch size $B=100$.
  (a) R-SGD compared with SGDM and NAG with adaptive momentum parameter  $\rho_t=\Gamma(t)$ (a smooth averaged version of R-SGD). (b) R-SGD compared with prefixed-momentum-parameter SGDM.
  (c) R-SGD compared with Adam and adaptive-momentum-parameter SGDM ($\rho_t=\Gamma(t)$). We use the same reinforcement schedule as that used in Sec.~\ref{res-toy} for R-SGD.
  Deep learning is performed with ReLU activation ($f(x)={\rm max}(0,x)$) and cross-entropy cost function. The classification error is measured as the fraction of the testing examples which the trained neural network mis-classifies.
  }\label{sgdm}
\end{figure}

Although the reinforcement probability specified by ($\gamma_0,\lambda$) we used in the synthetic dataset works on the MNIST dataset (see an example in Fig.~\ref{sgdm} (c)), we found that the reinforcement probability $\Gamma(t)=1-1/\sqrt{t+1}$ works better (shortening the training time to reach a lower generalization error) for MNIST dataset, and furthermore, this choice does not saturate
the reinforcement probability to one within the explored range of learning time (Fig.~\ref{powerlaw}), which offers a nice candidate to demonstrate the performance difference between R-SGD and SGDM.
Fig.~\ref{mnist} shows that R-SGD improves significantly over BackProp, reaching
a similar test performance to that of Adam with moderate training data size ($M=10K$). R-SGD achieves a test error of $0.0535\pm0.0021$, compared with BackProp reaching
$0.1047\pm0.0020$, and Adam reaching $0.0535\pm0.0016$ (as in Table~\ref{tab}). The test error is averaged over five independent runs (different sets of training and test examples).
Note that, as training size increases (e.g., $M=15K$), the test performance of R-SGD becomes slightly better than that of Adam (Table~\ref{tab}). Adaptive methods such as Adam often show faster initial progress, but their performances get quickly trapped by a test error plateau~\cite{wilson-2017}. In contrast, as shown in the inset of Fig.~\ref{mnist}, R-SGD as a non-adaptive method is able to 
reach a lower test error by taking a few more epochs (note that Adam needs more computer memory to store the uncentered variance of the gradients).

\begin{table}
\caption{Test error of different methods on MNIST dataset with $2000$ testing images and different training data sizes ($10000$ and $15000$). The test error is averaged
over five independent runs with $150$ epochs.}
\label{tab}
 \begin{tabular}{ccc}
 \hline\hline
\multirow{2}{*}{Methods} &
      \multicolumn{2}{c}{Test error}\\ 
      \cline{2-3} 
      & {10K} & {15K} \\
      \hline
 BackProp    & $0.1047\pm0.0020$ & $0.0948\pm0.0030$ \\
R-SGD & $0.0535\pm0.0021$ & $0.0465\pm0.0018$\\
Adam    & $0.0535\pm0.0016$ & $0.0471\pm0.0022$\\
\hline\hline
\end{tabular}
\end{table}
  
  As shown in Fig.~\ref{sgdm} (a), SGDM with adaptive momentum parameter reaches a higher test error than R-SGD, which confirms that the switch-like event plays an
  important role in guiding R-SGD to a good region. In R-SGD, the reinforcement along some weight components is turned off 
  with a finite probability, then along these directions, only the current gradients are used (the same as those used in BackProp). But the reinforcement may be turned on along 
  these directions once again during training. In contrast, for SGDM, the momentum term with adaptive momentum parameter is always applied to update 
  all weight components during training. This mechanism difference leads to different test performances observed in Fig.~\ref{sgdm} (a). The training dataset may change the error surface in practice. It seems that
  the performance of R-SGD is robust to this change (Table~\ref{tab}). NAG with adaptive momentum parameter $\rho_t=\Gamma(t)$ learns quickly but gets trapped by a slightly higher test error. In addition,
  a single epoch in NAG takes an extra computational cost of $\frac{M}{B}|\mathbf{W}|$ due to the partial update.
  
  For fixed-momentum-parameter SGDM ($\rho_t$ does not change over time, unlike the reinforcement probability in R-SGD), the learning performance gets worse (Fig.~\ref{sgdm} (b)).
  When applied to a deep network with ReLU activation and cross-entropy as the objective function, R-SGD still shows competitive performance (Fig.~\ref{sgdm} (c)).

\begin{figure}
\centering
 \includegraphics[bb=0 0 819 572,scale=0.3]{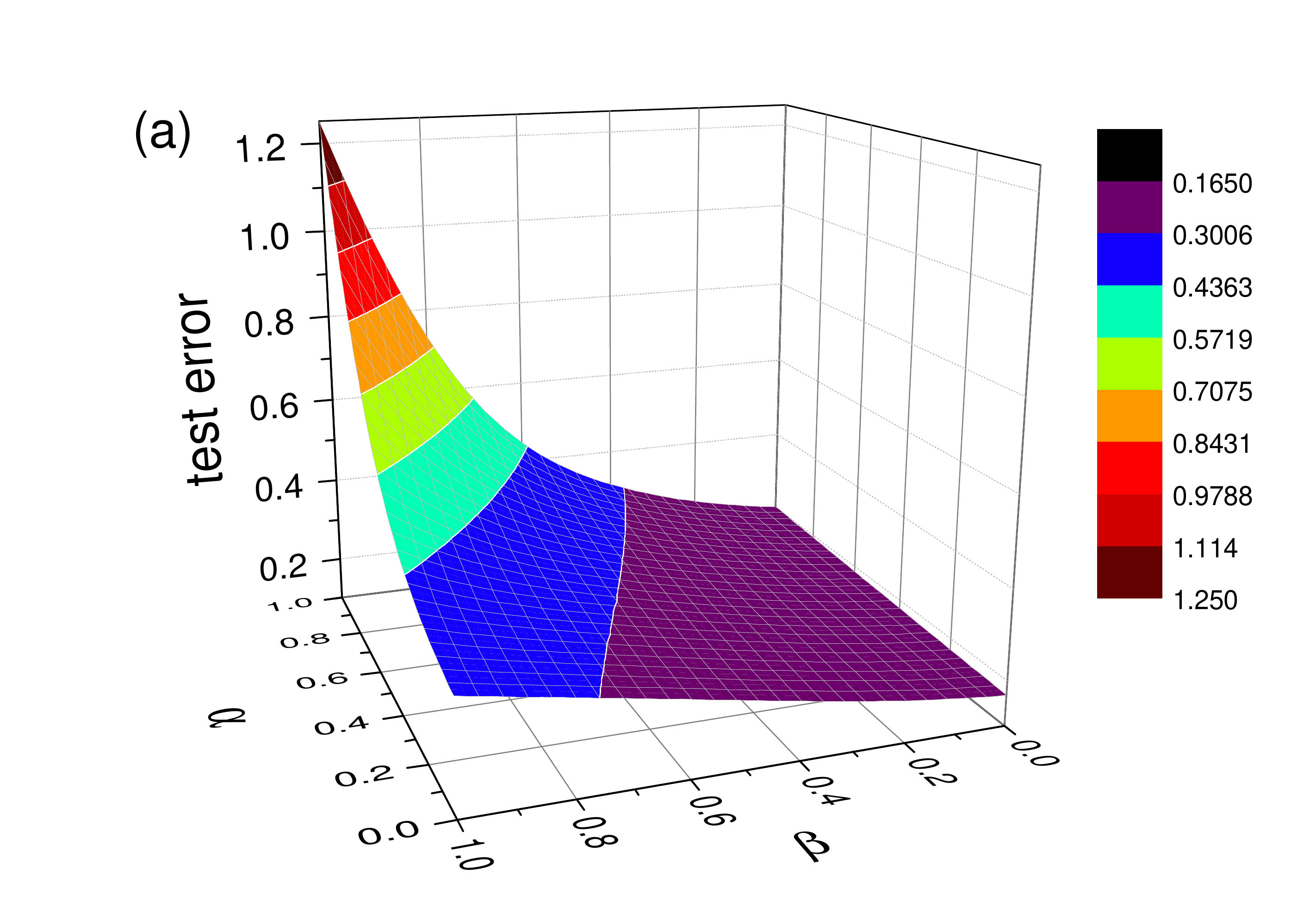}
     \hskip .05cm
  \includegraphics[bb=0 0 792 612,scale=0.3]{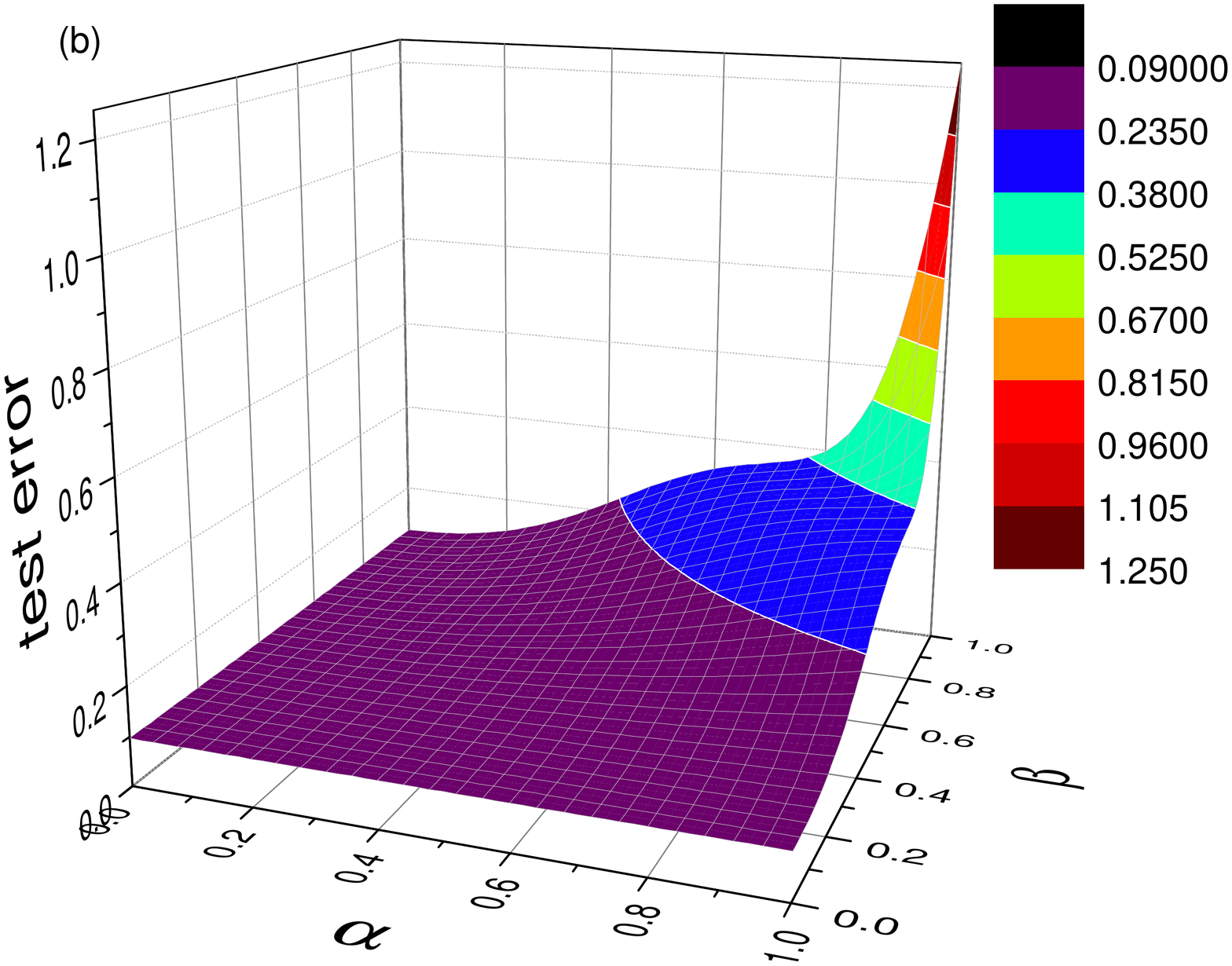}
     \vskip .05cm
     \includegraphics[bb=0 0 819 572,scale=0.3]{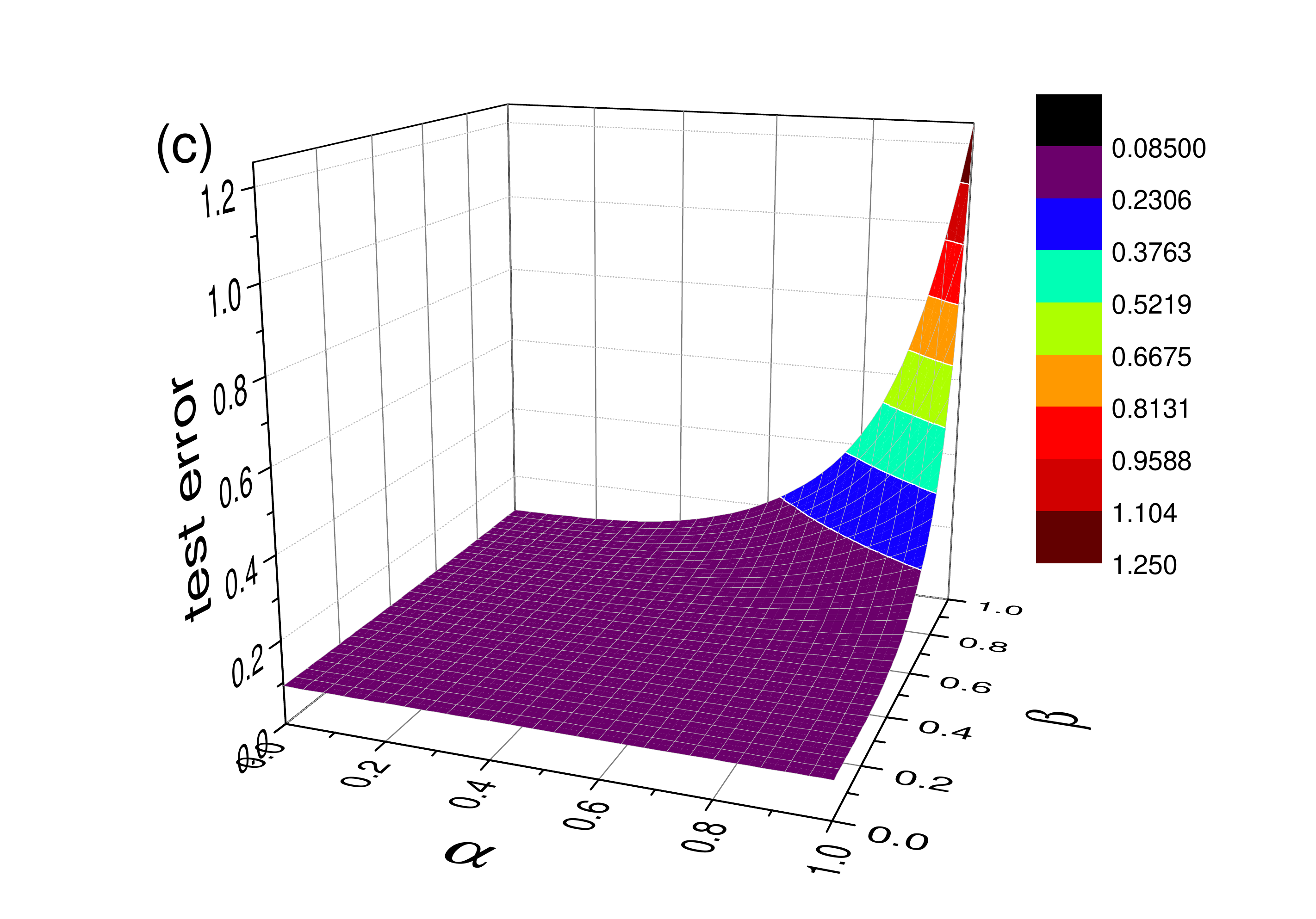}
     \vskip .05cm
  \caption{(Color online) Bilinear interpolation of weight configurations selected from one learning trajectory with corresponding epochs $t_{ep}=0,30,60,80$. The network is trained on
  $M=1000$ examples with the other $1000$ examples as test data. $B=100$. (a) BackProp. (b) R-SGD. (c) Adam.
     }\label{topoTrj}
 \end{figure}

\begin{figure}
\centering
 \includegraphics[bb=0 0 819 572,scale=0.3]{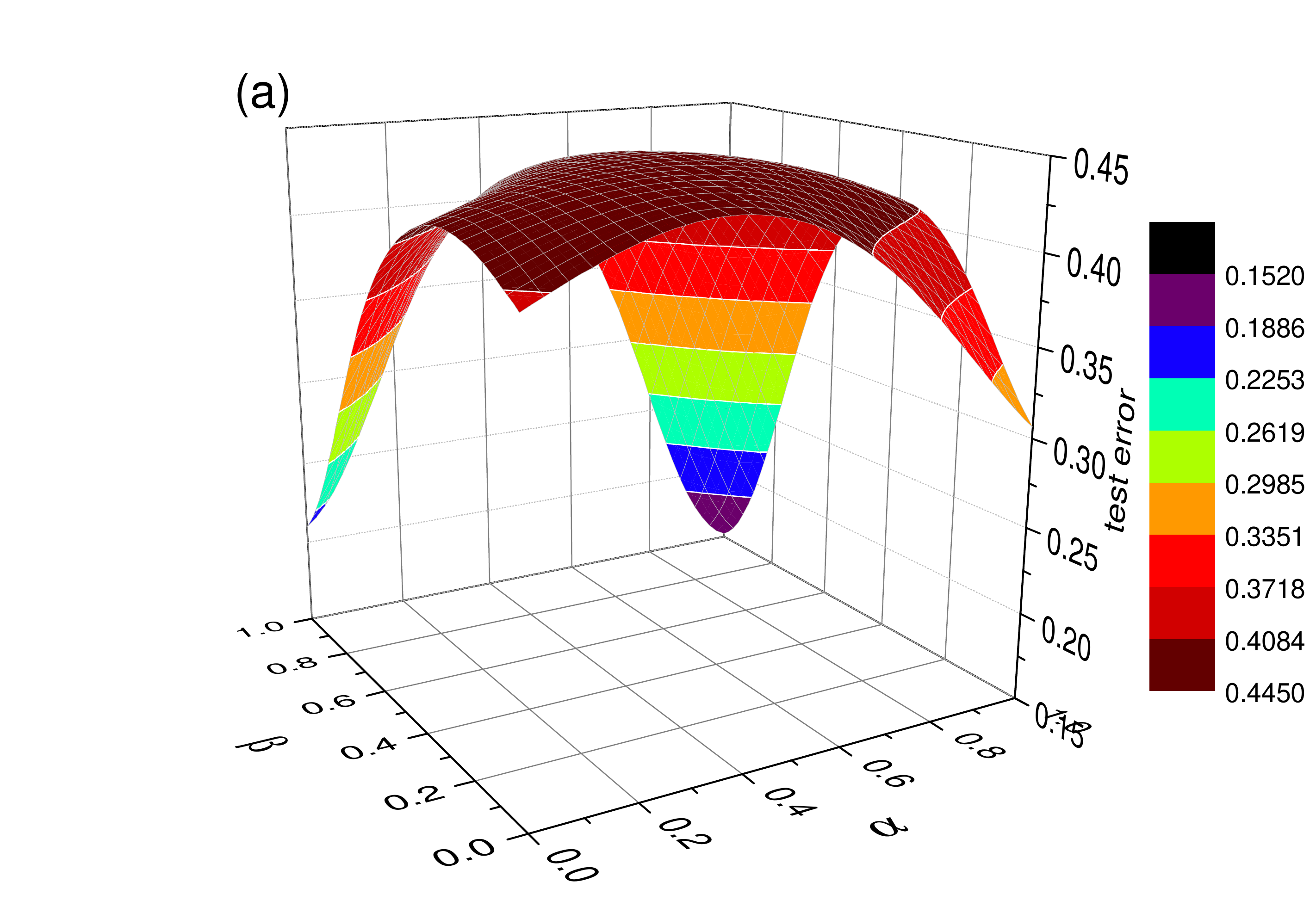}
     \hskip .05cm
  \includegraphics[bb=0 0 792 612,scale=0.3]{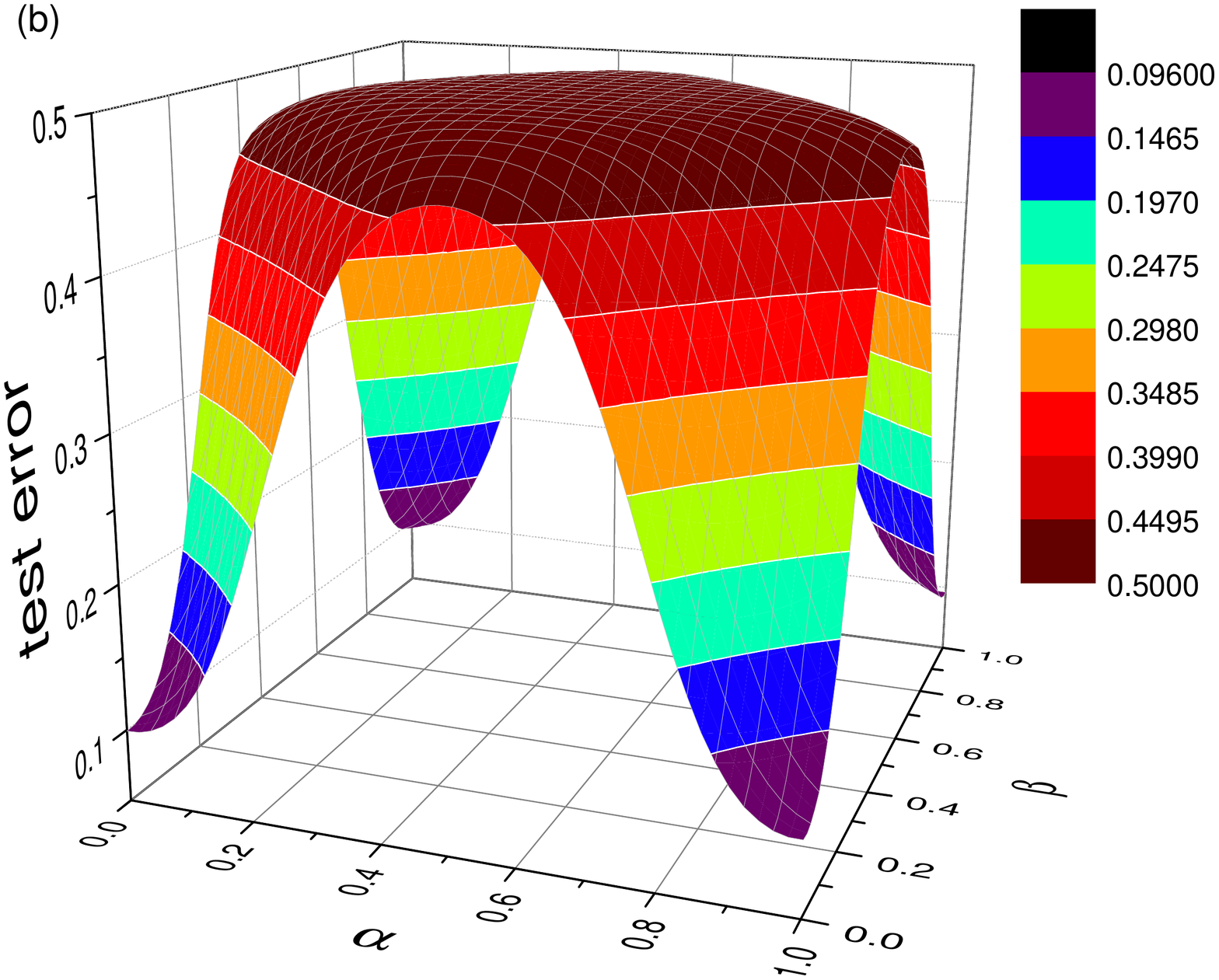}
     \vskip .05cm
     \includegraphics[bb=0 0 819 572,scale=0.3]{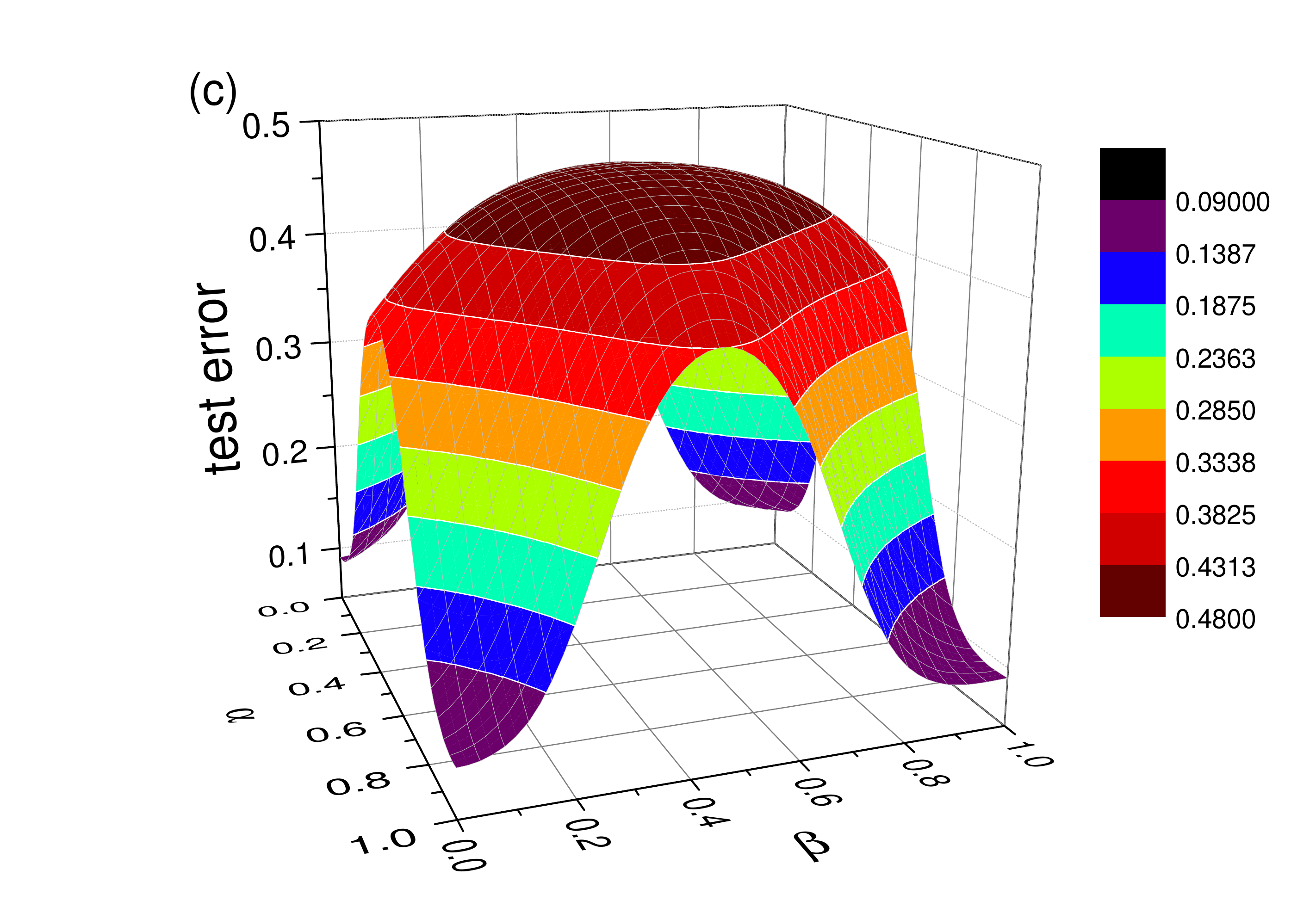}
     \vskip .05cm
  \caption{(Color online) Bilinear interpolation of four different solutions obtained starting from four different random initializations. The network is trained on
  $M=1000$ examples with the other $1000$ examples as test data. $B=100$. (a) BackProp. (b) R-SGD. (c) Adam.
     }\label{topoSol}
 \end{figure}

We also use bilinear interpolation method to qualitatively analyze the test error surface of the three algorithms (BackProp, R-SGD and Adam). Using the bilinear interpolation method~\cite{Im-2016}, one can
visualize the error surface in $3$D subspace spanning four high-dimensional weight configurations. These four weight configurations defined as $\{\mathbf{W}_i\}_{i=1}^{4}$ are chosen either from
one learning trajectory or from solutions obtained starting from four different random initializations. Based on these four configurations, the error function is varied as a function of a new
constructed weight matrix specified by $\mathbf{W}=\beta(\alpha\mathbf{W}_1+(1-\alpha)\mathbf{W}_2))+(1-\beta)(\alpha\mathbf{W}_3+(1-\alpha)\mathbf{W}_4)$, where $\alpha\in[0,1]$ and $\beta\in[0,1]$ are two control parameters.
Results are shown in Fig.~\ref{topoTrj} for one trajectory interpolation. BackProp decreases the test error slowly, and finally reaches a plateau and get stuck there.
Adam decreases the error very quickly, and reaches a more apparent and lower plateau than that of BackProp. Remarkably, R-SGD first decreases the error quickly, which is followed by a plateau. The
plateau is then passed and finally R-SGD reaches a region with less apparent flatness. Interpolation results of four different solutions are shown in Fig.~\ref{topoSol}.
Clearly, BackProp can get stuck by different solutions of different qualities, depending on initializations. Both Adam and R-SGD can reach solutions of nearly the same
quality, despite different initializations. However, the subspace looks sharper for R-SGD than for Adam. This rough analysis by using bilinear interpolation
method implies that the high dimensional topology of weight space seen by the three algorithms might be intrinsically different. However, to characterize necessary properties of a good region with nice
generalization capabilities requires a theoretical understanding of the entire high dimensional weight space in terms of analyzing some non-local quantity. Establishing a theoretical 
relationship between learning performances of various SGD algorithms and the intrinsic structure of the error surface is still an extremely challenging task in future studies.

\section{Concluding remarks}
\label{conc}
  In this paper, we propose a new type of effective strategy to guide SGD to overcome the plateau problem typical in deep neural network learning.
  This strategy takes into account previous (accumulated) gradient information in a probabilistic way when updating current weight matrix. It introduces a stochastic reinforcement to
  current gradients, and thus enhances the exploration ability of the original SGD. This strategy is essentially different from an independently random noise added to the gradient during
  the learning process~\cite{Neel-2015,Chau-2015}. In fact, we add time-dependent Gaussian noise to gradients during training in our simulations using default hyper-parameters~\cite{Neel-2015}, whose performance could
  not be comparable to that of R-SGD within $100$-epochs training (or it requires longer convergence time). 
  
  A similar reinforcement strategy has been used in a single layer neural network with discrete weights~\cite{Zecchina-2006}, where local fields in a belief propagation equation are
  reinforced. In our work, we study deep neural networks with continuous weights, and thus the gradients are reinforced. 
 The reinforced belief propagation is conjectured to be
  related to local entropy maximization~\cite{Baldassi-2015,Baldassi-2016} in discrete neural networks. Whether R-SGD reshapes the original non-convex error surface in some way, such that searching for
  a parameter region of good-quality is facilitated, remains an interesting open question. We leave this for future work. 
  
  In the current setting, the learning performance of R-SGD is comparable to (in MNIST) or even better than that of Adam (in the synthetic dataset), which requires one-fold more computer memory to store the uncentered variance of
  the gradients. The learning step-size of Adam is adaptively changed, which means that the step-size is automatically tuned to be closer to zero when
  there is greater uncertainty about the direction of the true gradient~\cite{Adam}, while R-SGD uses the stochastic reinforcement of the gradient information to
  deal with this kind of uncertainty, and shows comparable and even better performance.
  
  A recent study argued that adaptive learning methods such as Adam generalize worse than SGD or SGDM on CIFAR-10 image classification tasks~\cite{wilson-2017}.
  It is thus very interesting to evaluate the performance of R-SGD on more complicated deep network model and complex datasets, 
  which are left for future systematic studies.
  
  R-SGD may be 
  able to avoid vanishing or exploding gradient problem typical in training a very deep network~\cite{Bengio-2010}, probably thanks to accumulation of gradients used stochastically. 
  In addition, it may take effect in recurrent neural network training~\cite{Ben-2012}.
  In fact, previous gradient information at each step can be
  weighted before accumulation according to its importance in guiding SGD. This is a very interesting direction for future studies on fundamental properties of R-SGD.



\begin{acknowledgments}
We are grateful to the anonymous referee for many constructive comments.  H.H. thanks Dr. Alireza Goudarzi for a lunch discussion which later triggered the idea of this work. This work was supported by the program for Brain Mapping by Integrated Neurotechnologies
for Disease Studies (Brain/MINDS) from Japan Agency for Medical Research and development, AMED, and by RIKEN Brain Science Institute.
\end{acknowledgments}


\end{document}